# ESLO: from transcription to speakers' personal information annotation

**Iris Eshkol[1], Denis Maurel[2], Nathalie Friburger[2]**

[1]Université d'Orléans, LLL
[2]Université François Rabelais Tours, LI
France
E-mail: iris.eshkol@univ-orleans.fr, {denis.maurel, nathalie.friburger}@univ-tours.fr

**Abstract**

This paper presents the preliminary works to put online a French oral corpus and its transcription. This corpus is the Socio-Linguistic Survey in Orleans, realized in 1968. First, we numerized the corpus, then we handwritten transcribed it with the Transcriber software adding different tags about speakers, time, noise, etc. Each document (audio file and XML file of the transcription) was described by a set of metadata stored in an XML format to allow an easy consultation. Second, we added different levels of annotations, recognition of named entities and annotation of personal information about speakers. This two annotation tasks used the CasSys system of transducer cascades. We used and modified a first cascade to recognize named entities. Then we built a second cascade to annotate the designating entities, i.e. information about the speaker. These second cascade parsed the named entity annotated corpus. The objective is to locate information about the speaker and, also, what kind of information can designate him/her. These two cascades was evaluated with precision and recall measures.

## 1. Corpus presentation

The most extensive examination of spoken French before 1980 is the Socio-Linguistic Survey in Orleans (*Enquête Socio-Linguistique à Orléans*, ESLO 1). This investigation was carried out towards the end of the Sixties by British academics with a didactic aim and represents a collection of 200 interviews with several references (sociological characterization of the interviewed persons, identification of the interviewer, date and place of the interview). The interview was recorded in a professional or private context, in a total of 300 hours of speech and the corpus contains approximately 4,500,000 words[1].

Note that a new survey, ESLO 2, has been undertaken by the LLL-Orléans in order to constitute a corpus comparable in terms of data gathering and archiving. ESLO 1 and ESLO 2 will form a collection of 700 hours of recording, that is more than 10 000 000 of words.

In this article we want to show the different levels of corpus annotations. Starting from the transcription which, we believe, is the first level of annotation, we will then describe how to increase the transcription with some semantic annotations in XML format.

## 2. Preliminary work on ESLO

The ESLO project contains many steps: numerization, coding of metadata, transcription synchronic with the sound, annotation, anonymization, tools of request and diffusion. The main theoretical and technical choices operated during the scientific exploitation of the corpus ESLO1 answer to a precise objective: to participate in the reflection on the evolution of models and methods of constitution and exploitation of the oral corpus for linguistic purposes [Abouda, Baude, 2007].

ESLO1 was consisted of the wave band of recordings, index card of sociological characteristics, descriptions of situation (date, place, remarks on the acoustic, etc.). Audio files were digitized; an indexation and a first cataloguing were accomplished. The objective is to make available all data to the scientific community in a format which allows optimal and intensive exploitation [Baude, 2006].

### 2.1 Transcription

We consider transcription to be the first stage of annotation because it is an example of the important particular operation of oral corpus encoding. The phase of transcription was complicated in particular by the lack of normalized practices within the scientific community. After an expertise of the actual use of tools and conventions of transcription, the team developed proposals of transcription based on the following principles: simplifying navigation within the signal, transcribing close to norm but including the truncations, and others disfluences, avoiding typographical marks [Blanche-Benveniste, Jeanjean, 1987] and making a transcription of the complete corpus available as soon as possible without the implication of a specific linguistic theory and any analysis. It is simply a tool of corpus preparation. Therefore, the conventions of transcription were reduced to a minimum:

- the transcription is orthographic,
- the segmentation is made according to an intuitive unit of the type « group of breath » and/or appropriate syntactic unit;
- the turn-taking was defined by speaker's changes;
- the pauses were indicated automatically by their length (precision of a hundredth of a second);
- no punctuation except the points of exclamation and question marks;
- no uppercase letters except the named entities;

---

[1] The part of ESLO1 was put online (http://bach.arts.kuleuven.be/elicop/) within the Elicop project [Mertens, 2002].

- word truncation is indicated by the dash (mont-)

The rest is encoded with the automatic tags marked by the software used for transcription, i.e. Transcriber (turns of speech, acoustic conditions, tags of events like noises, lexicon, pronunciation, language, etc.). We chose this software for its simplicity of use, its robustness when treating long files, and its exit in a format XML which seemed us to be a guarantee of interoperability. The transcription made is synchronized with sound. This is an example of transcription XML file:

<Turn speaker="spk4" startTime="10.88" endTime="14.843">
<Sync time="10.88"/>**vous vous plaisez à Orléans?**
<Sync time="12.721"/></Turn>

(*Do you like it at Orléans?*)

## 2.2 Cataloguing

We transformed the audio recording to audio files (wav format), the handwritten transcription became transcription files XML of Transcriber. Each document (audio file and XML file of the transcription) is described by a set of metadata stored in an XML format. The data base php/mysql with the information about the sound recording, the speakers and the questionnaire taken from the original card index was created. An interface (xquery) was accomplished within the framework of GRICO[2] and CRDO[3] project by the specialists in the management of the oral corpus and the linguistic researchers of LLL (Figure 1, Annex 1)

The data base ESLO includes eight tables:

- Recording files (type of recording: interview or other, date, place, etc.);
- Speakers (date and place of birth, profession, age, sex, etc.);
- Questionnaire;
- Transcriptions (number of recording, names of transcribers, date of transcription);
- Table which allows the relation between the recording files and questionnaire (problems of sound, of conventions, etc.)
- Members of the team;
- Set of problems met by the transcribers,
- Remarks marked by the transcribers (syntactic structure, complementary questions, etc.)

The data base allows the consultations of ESLO1 by the list of recordings, transcriptions and speakers (Figure 2, Annex 1). So, it is possible to look the description of each recording (type, during, place, date, acoustics, etc.) with the link to speakers characteristics, the representation of the transcriptions (name of transcribers, type of transcription: brut, read or validated, problems and remarks of transcribers about transcription) with the possibility to submit a new transcription, the characteristics of the speakers (date and place of birth, sex, profession, etc.) with the link to recordings. The new interface of database is in the process of creation today. It will allow the consultation more detailed of our corpus: to extract the transcription with recording according to string search, using sociological characteristics of speakers, questions answered etc.

## 3. Personal information annotation

After this preliminary work, we wanted to add personal information annotation about the different speakers, always in XML format. But the progress of the transcription make us use only a part of the corpus, 112 Transcriber XML files (32 577 Kb), i.e. 120 hours of speech. We split it into 105 work files (31 004 Kb) and 7 evaluation files (1 573 Kb), i.e. 5.1% of the corpus.

To insert annotation tags, we used hand built local grammars, in the form of transducer cascades [Abney, 1996] and we parsed the corpus with the CasSys system [Friburger, Maurel, 2004], based on Unitex software [Paumier, 2003]. Entity recognition may be need the succession of two or more transducers, in a specific order. The parsing preprocesses the file, segments it using XML Transcriber tags[4] [Dister, 2007].

In fact, we used two cascades; the first one for named entity recognition and the second one for discovering speaker information. The second cascade used the first one.

### 3.1 First cascade: named entity recognition

We adapted a precedent work realized some years ago for a newspaper corpus and we adapted it to our corpus. We used eight types[5]: person (*pers*), post (*fonc*), organization (*org*), location (*loc*), product (*prod*), date and hour (*time*), amount (*amount*) and event (*event*). Each type is subdivided in subtypes, for instance human (*pers.hum*), political post (*fonc.pol*), facilities (*loc.fac*), etc. See [Maurel and al., 2009] for the complete list.

For instance, we tagged:

moi je suis native de <EN type="loc.admi">**Pithiviers**</EN> j'aime mieux <EN type="loc.admi">**Orléans**</EN>

(*I was born at the city of Pithiviers, I prefer the city of Orléans*)

Figure 3, Annex 2, presents the transducer recognizing the context of "musician", for instance *Le musicien Willy DeVille*, or

*le concert de Johnny Hallyday.*

The evaluation we computed for the first cascade is presented Table 1. See [Maurel and al., 2009] for more information about these results.

---

[2] Groupe de Recherche sur l'Interopérabilité des Corpus Oraux. Michel Jacobson (Lacito-CRDO-DAV) et Richard Walter (IRHT).
[3] Centre de ressources pour la description de l'oral http://crdo.vjf.cnrs.fr:8080/exist/crdo/
[4] For written corpora, this segmentation usually uses sentence boundary detection [Friburger and al., 2000].
[5] The seven first types were defined for the ESTER campaign about French spoken language transcription.

|  | Precision | Recall |
|---|---|---|
| Entities | 97.8% | 94.0% |
| Entity types | 92.0% | 88.4% |
| Entity brackets | 91.1% | 87.5% |

Table 1: Evaluation of the first cascade

### 3.2 Second cascade: designating entity recognition

The second cascade parsed the named entity tagged corpus. The objective is to locate information about the speaker and, also, what kind of information can designate him/her. We call this information designating entities [Eshkol, 2009].

The survey focused on the speaker and his/her family. The cascade inserted two different tags: one for the person who speaks or whom one speaks and the other one for the information about him/her.

The concerned persons were the speaker (*pers.speaker*), his/her spouse (*pers.spouse*), his/her children (*pers.child*) and his/her parents (*pers.parent*).

A lot of questions dealt with personal information, as his/her age (*identity.age*), his/her birth date (*identity.birth*) and where he/her came from (*identity.origin*), when he/her came at Orléans (*identity.arrival*) and also information about his/her children (*identity.children*), etc. Another group of questions dealt with his/her work or the work of his/her spouse or children (*work.occupation*, *work.field*, *work.location*, *work.business*), etc.

For instance, we tagged:

<DE type="pers.speaker"> **moi je suis** <DE type="identity.origin"> **native de** <EN type="loc.admi"> **Pithiviers**</EN> </DE> </DE>

(*I was born at the city of Pithiviers*)

This second example is more complex, with three different kinds of tag: transcription, named entity and designating entity:

**alors ça fait longtemps que vous habitez** <Event desc="pi" type="pronounce" extent="instantaneous"/> **euh** <NE type="loc.admi"> **Orléans** </NE> **hein?** </Turn> <Turn speaker="spk1" startTime="16.224" endTime="18.444"> <Sync time="16.224"/> <Sync time="17.338"/> <DE type="pers.speaker"> <DE type="identity.arrival"> <NE type="time.date.rel"> **neuf ans** </NE> </DE> </DE>

(*Then you leave at Orléans for a long time?/nine years*)

A third example deals with the family of the speaker:

<DE type="pers.speaker"> **nous sommes revenus parce que** <DE type="pers.parent"> **mon père était** <DE type="work.field"> **officier** </DE> </DE> </DE>

(*We came back because my father was officer*)

Figure 4, Annex 2, presents the transducer recognizing the context of speaker identity, for instance *I am born at Orléans*, *I left at Paris*…:

The evaluation we computed for the second cascade is presented Table 2, but the results are open to doubt because information is very sparse. There are just 77 designating entities in the seven test files…

| Precision | 94.2% |
|---|---|
| Recall | 84.4% |

Table 2: Evaluation of the second cascade

## 4. Conclusion

This paper presents three years work on a survey corpus from 1968. It deals with transcriptions, cataloguing of data and others annotations concerning named and designating entities that were added to. The transcription used Transcriber software and the entities parsing used CasSys, a transducer cascade tool. The objective is to make available all data to the scientific community in a format which allows its optimal and intensive exploitation. One of the possible applications is navigation, visualization/listening and extraction of some (semantic, sociolinguistic, etc.) information using the corpus itself (transcription/sound files), its annotations and the data base with the metadata describing the corpus. So, this project shows how the computer technology is used to emphasize and to explore linguistic variations.

### Acknowledgements

This work was supported by the French ANR project *Variling* and the Feder (European Regional Development Fund) project *Entités*.

## Annex 1 Cataloguing

Figure 1: Characteristics of speaker

Figure 2: Homepage of ESLO1 consultation

# Annex 2 Personal information annotation

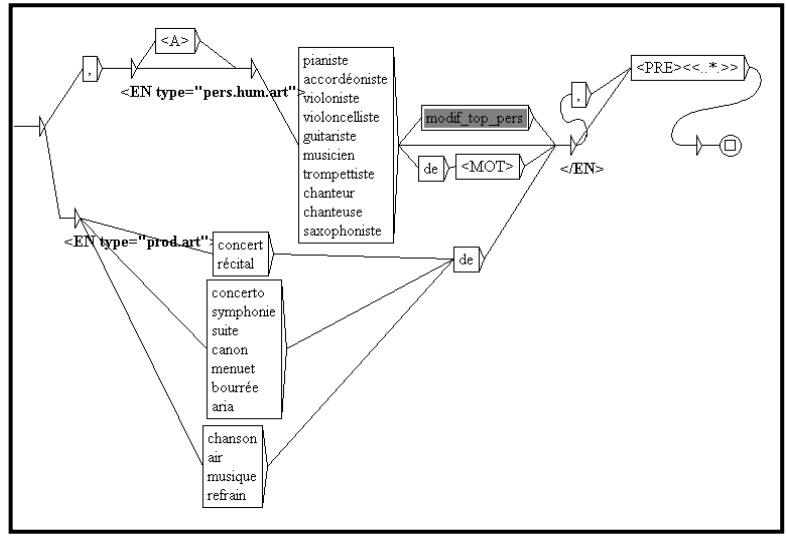

Figure 3: Context of "musician"

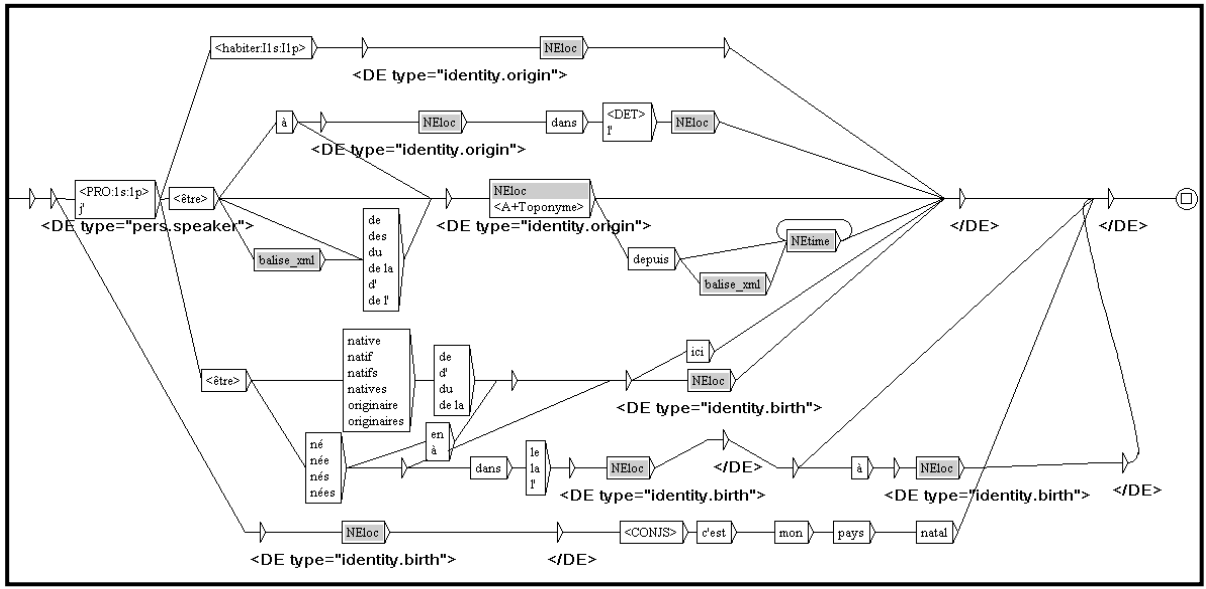

Figure 4: Context of speaker identity